# Free Space Estimation using Occupancy Grids and Dynamic Object Detection

Raghavender Sahdev
*Department of Computer Science*
*University of Toronto*
*Toronto, Canada*
raghavender.sahdev@mail.utoronto.ca

*Abstract*—In this paper we present an approach to estimate Free Space from a Stereo image pair using stochastic occupancy grids. We do this in the domain of autonomous driving on the famous benchmark dataset KITTI. Later based on the generated occupancy grid we match 2 image sequences to compute the top view representation of the map. We do this to map the environment. We compute a transformation between the occupancy grids of two successive images and use it to compute the top view map. Two issues need to be addressed for mapping are discussed - computing a map and dealing with dynamic objects for computing the map. Dynamic Objects are detected in successive images based on an idea similar to tracking of foreground objects from the background objects based on motion flow. A novel RANSAC based segmentation approach has been proposed here to address this issue.

*Keywords-Free Space; occupancy grids; dynamic programming; disparity; RANSAC,*

## I. INTRODUCTION

Autonomous Driving has been researched a lot recently. One of the major tasks of driving an autonomous vehicle is to find the drivable collision free space for it to drive on. For the most part this is often restricted to roads in an urban environment. Free Space in the context of autonomous driving is defined as the collision free space in which the car can freely navigate which means it has no obstacles (other cars, humans, buildings, trees, pavement, etc.). Finding free space implies getting rid of all other obstacles apart from the space where the car can drive without collisions which is the road in most cases. Here we present an approach to compute free space from a stereo image pair using probabilistic occupancy grids. The occupancy likelihood for each of the grids is estimated and finally we use a thresholding based segmentation to compute the free space. Occupancy grids were first introduced by Elfes in 1989 [1]. A 2D occupancy grid refers to a grid representation of the top view of the map as seen from a bird's eye. It models the world by using a probabilistic tessellated representation of spatial information which Elfes called the Occupancy grid. Occupancy grids have been successfully used to build maps and some systems use them to plan collision free paths, identify environmental features for localization, collision avoidance, human detection, etc. In this paper we use occupancy grids to model the world environment and estimate free space in an urban environment from the KITTI dataset [21].

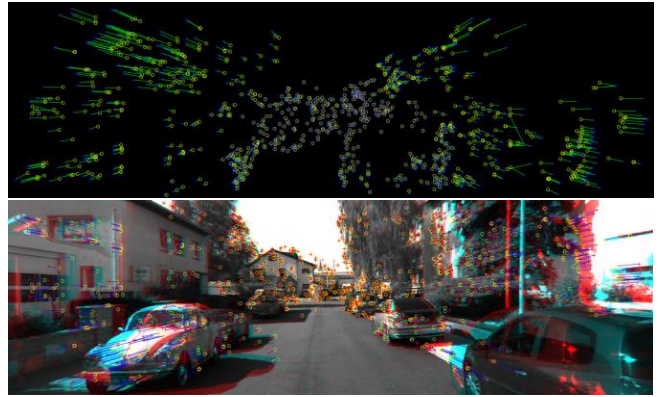

Figure 1. Perspective Motion as seen from a car; Flow vectors (top). Points closer to the car have greater flow than far away points. Background points follow a model for the flow.

After computing the free space from a single image we generate a mapping of the environment by fusing two occupancy grid maps obtained from successive images. We fuse the occupancy grid maps by computing a transformation between the successive occupancy grid maps w.r.t. a global coordinate system. One important issue that needs to be handled here is the problem of having dynamic objects in the scene. Dynamic objects are essentially moving obstacles and do not belong to the background. We detect dynamic objects by segmenting their motion from the background motion in a RANSAC manner. Intuitively the flow of the background should move in a way that is different from the foreground motion. For instance for a camera mounted on car travelling straight would lead to a perspective motion as shown in Figure 1. So all back ground points would move backwards in such a scenario and obey a model. The points not obeying this model are the dynamic objects. Once the dynamic objects are detected we find their position in the 2D occupancy grid map and do not use it as an occupied spot during the update process of the map and treat it as unknown. We make a simple transformation of the points of detection of dynamic objects from image to the occupancy grid which contains the object. Finally we validate our proposed methodology on the KITTI benchmark dataset [21].

The major contributions of this paper can be summarized as follows:
- Computing free space from an image based on occupancy likelihood estimations using a thresholding based segmentation approach
- Detecting dynamic objects in the scene using a novel RANSAC based segmentation of dynamic objects and the background is proposed by us. We then integrate the information obtained from the detection and use it for mapping.
- Mapping the free space as seen from a bird's eye view and building a map based on occupancy grids is discussed briefly.

The paper is organized as follows. In section II we present the relevant work carried out to estimate free space and detection of dynamic objects in a scene. Our proposed approach for computing free space is explained in section III. Section IV presents the approach for handling dynamic objects and mapping. We present the experimental evaluation of our method in section V. Section VI has the implementation of the approach and finally present conclusions and directions for possible future work in section VII.

## II. Relevant Work

In this section we review the work done for estimation of free space and detection of dynamic objects in a scene. First we present the relevant work done for computing free space and later in this section describe relevant work for detection of dynamic objects in a scene. Here we use the information obtained from these two components (estimating free space and dynamic object detection) and propose an approach to build a map of the environment which gives an occupancy map representation on the scene. Free Space estimation has been tackled by numerous researchers by approaches like using occupancy grids and appearance based approaches using visual cues from the image. First we present the relevant literature which focuses on Occupancy grid based approaches and later we present work which majorly uses visual cues to segment out free space from the obstacles present in a scene. Occupancy Grids were proposed by Elfes [1] as mentioned previously. He used a sonar sensor to build a map of the environment using the concept of occupancy likelihood of a grid. The likelihood of each grid was modeled stochastically and the map was built considering the position uncertainty of the robot and the robot view. Sonar sensors in general are more precise than vision sensors and are accurate enough to detect fine details. In this paper we rely on vision to compute occupancy grids by computing disparities. The perfect quality of disparity (not possible realistically) is analogous to the high quality of a sonar sensor. Getting good disparity should theoretically achieve results as accurate as a sonar sensor. Occupancy Grids have been used to solve the mapping problem by approaches like the Inverse Sensor Model [2] and Forward Sensor Models [3]. The key difference between the inverse sensor and forward sensor model is that inverse sensor models do not take occupancy of the neighboring cells into account for predicting the occupancy likelihood of the current cell whereas in the forward sensor model proposed by [3] takes neighboring cells also into account. The forward sensor model uses the Expectation Maximization algorithm for searching the maps and finding the most likely map.

Some of the early work for free space detection includes that of Chrisman and Thorpe (1991) [19] where they used color information to detect road for robot navigation. In 1994, Zhang and Nagel [20] presented an approach to compute texture orientation fields based on a covariance matrix obtained from the gray scale image. Then they do unsupervised segmentation of roads in the initial phase followed by a supervised segmentation. Labayrade et al. [4] do real time obstacle detection using a v-disparity image; the v-disparity image provides a good and robust representation of the geometric content of the road scene; their proposed approach not only works on flat roads but also on non-flat geometric surfaces. Another famous work is that of Oniga and Nedevschi [5]. They use the concept of a digital elevation map obtained from 3D from a stereo camera. RANSAC technique is used to fit a quadratic road surface to the region in front of the camera. Finally they classify each of the cells obtained from the digital elevation map to be one of road, traffic isle or an obstacle based on road surface. Wedel et al. [6] find free space by extending the v-disparity approach by modeling the road by fitting a B-Spline curve and estimating the parameters by means of a Kalman Filter. Lee et al. [7] estimate free space by using the longitudinal profile of the road surface to classify each disparity map into an obstacle disparity map or a road surface disparity map. On these disparity maps they use dynamic programming to estimate a border line separating the road surface and the non-road surface on a $u$-disparity representation. Maier and Bennewitz [8] estimate the ground plane from monocular images by using appearance based features and odometry estimates; they validate their approach by making a humanoid robot traverse in a controlled structured environment. Their approach iteratively estimates ground plane based on detected flow. They use floor features and train visual classifiers to label the traversable area in successive camera images. Some other approaches for estimating free space include [9] and [10]. In [9], Miranda et al. estimate drivable image area from monocular images and in [10] Soquet et al. present an enhanced v-disparity algorithm to estimate free space. Hautière et al. [17] detect free space in foggy weather conditions based on restoring the road surface based on the fog density. Yoshizaki [18] estimate free space by detecting the dominant plane from a sequence of omnidirectional images obtained from a camera mounted on an autonomous robot. Fazl-Ersi and Tsotsos [37] propose a stereo based technique for detecting floor and obstacle regions in indoor environments, assuming the robot moves on a planar surface. Their approach does pixel classification at a higher level where neighbor pixels with similar visual properties are processed and classified together. After computing the corresponding points in the left and right images, for each pair of matched points, its symmetric transfer distance from the ground homography is computed. Regions with enough points to the ground plane homography are classified as ground and rest as obstacles.

Recently Yao et al. [14] proposed an approach to estimate free space from a monocular video sequence. They formulate this problem as one of inference in a Markov Random Field which estimates for each image column, the vertical coordinate of the pixel that separates the free space from obstacles. They formulate an energy function and minimize it to segment the image into two regions - free space and non-free space. This approach is in some sense similar to the dynamic programming part proposed by [11]. In [15] Badino et al. proposed a method to fit stixels (vertical lines) to the detected obstacles. They initially compute the disparity from a given stereo pair by using a variant of the semi global matching and mutual information method of [17] and use this disparity to compute 2D occupancy grids. The space found in front of the first obstacle is declared as free space. Finally they do height segmentation by computing a cost function and use dynamic programming to minimize the cost.

In this paper we present an approach similar to that presented in [11], [12] and [13]. In [11] Badnio et al. present an approach to compute free space using stochastic occupancy grids and present three different types of occupancy grids – Cartesian, column and polar occupancy grids. They use dynamic programming to compute the free space based on the occupancy likelihoods obtained from the polar occupancy grid. Similar to the approach proposed by Badino et al. [11], Høilund et al. [12], [13] also use stochastic occupancy grids and dynamic programing to estimate free space. However Høilund et al. use a Kalman filter to update their disparity images to get accurate disparities which in a way was indicated in [11] too. Another detailed work is that of Zou [16] in which he uses Bayesian occupancy grids, dynamic programming for segmentation and finally tests his approach in a variety of real world scenarios taken from a camera mounted on a car to validate his proposed methodology.

Now we describe some of the relevant work done for detection of dynamic objects. Zhang et al. [23] propose a method to detect and segment moving object from surveillance videos using an adaptive Gaussian mixture model with a support vector machine classifier to segment the foreground motion from the background. Wang et al. [24] track dynamic objects using a 2D lidar mounted on an autonomous car. Bugeau and Perez [25] proposed a method for detection and segmentation of moving objects in highly dynamic scenes. Sabzevari and Scaramuzza [26] segment out motion of moving objects from perspective views and conduct experiments of their proposed approach on the KITTI benchmark. They factorize the projective trajectory of key-points and exploit the epi-polar geometry of calibrated cameras to generate several hypotheses for motion segments. And then use those hypotheses to estimate multiple motions. Vidal et al. [27] leverage the geometric and algebraic properties of the fundamental matrix to segment out the motion of moving objects. Yang and Wang [35] address the problem of ego-moiton estimation and dynamic object detection using a multi-model RANSAC paradigm. Jung et al [36] proposed a method called Randomized Voting to do rigid body motion segmentation. They leverage the concepts of epi-polar geometry to segment the motion of moving objects. A score is computed using the distance between feature points and the corresponding epi-polar line and the final clustering of the motion is done based on this score. Yan and Pollefeys [38] proposed an algorithm to do articulated motion segmentation called RANSAC with priors. It does not need knowledge of number of articulated parts. Priors are derived from the spectral affinities between every pair of trajectories. Their approach also applies to independent motion. In this work we segment dynamic motion from the background motion based on a model learnt by RANSAC algorithm. Our proposed approach is robust and efficient. Its robustness is ensured by the RANSAC nature. The model is obeyed by the background points and moving objects objects are treated as outliers by the model. After finding out the dynamic objects in the scene we find their corresponding position on the occupancy grid representation and for mapping purposes treat that point as an unknown space.

III. OCCUPANCY GRIDS TO COMPUTE FREE SPACE

Occupancy Grids are used to model the 3D world in the form of cells. The world could either be modeled by 3D cells/voxels or 2D cells. Here we use 2D occupancy grids. Figure 1 gives a brief overview of our presented approach to compute free space from a pair of stereo images.

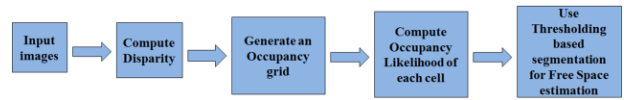

Figure 2. General pipeline used to compute free space in a single image.

### A. Disparity Computation

The first step in the proposed approach is to compute the disparity map from the given image stereo pair. In this paper we use the displets based approach proposed by Guney and Geiger [22]. This currently ranks best on the KITTI stereo dataset leader board. Their approach regularizes over large distances using object category specific disparity proposals. They do sematic segmentation to get the object shape and then using inverse graphics techniques know the structure of the model of the object. For this work we directly use their code to compute disparity.

### B. Occupancy Grid Representation

An occupancy grid is defined as 2D array which models the occupancy likelihood of the environment. Occupancy grid represents a tessellation of the real word into the form of grids/cells. The world is projected onto the ground plane and divided into cells of a specified resolution. Each cell corresponds to a specific area on ground. Two cells $C_{ij}$ and $C_{kl}$ may not have the same area. The indexes $i, k$ specify the horizontal lateral component and $j, l$ represent the depth in the real world. Each grid of the occupancy grid map represents an occupancy likelihood which indicates the probability of the grid being occupied or being free. A sample representation of an occupancy grid can be seen in Figure 2. Before computing the likelihood we introduce some notations as follows similar to [11]:

- $m_k$ is the $k^{th}$ measurement of the form $(u, v, d)^T$ where $u, v$ are the image coordinates of a pixel and $d$ is the computed disparity
- $m_k$ is the noisy measurement of some real but unknown vector $\bar{m}_k$
- $\delta_k = m_k - \bar{m}_k$ is real unknown error

We here assume that $\delta_k$ comes from a random Gaussian process with zero mean and the probability density function given as:

$$G_{m_k}(\delta_k) = \frac{1}{(2\pi)^{\frac{3}{2}}|\tau_k|} \exp -\frac{1}{2} \delta_k^T \tau_k^{-1} \delta_k \quad (1)$$

where $\tau_k$ is the real measurement covariance matrix. The function $G_{m_k}$ models the likelihood of obtaining an error $\delta_k$. Now we define $L_{ij}(m_k)$ which computes the occupancy likelihood estimate of the cell $i, j$ for the given measurement $m_k$. For a given cell we sum up all the likelihoods obtained from $m$ measurements to compute the occupancy likelihood over the cell $i, j$. We estimate the occupancy likelihood of each grid cell as:

$$D(i,j) = \sum_{k=1}^{m} L_{ij}(m_k) \quad (2)$$

where $L_{ij}$ is the likelihood function $G_{m_k}$ as defined in (1). In this work we use a column disparity grid as proposed in [11].

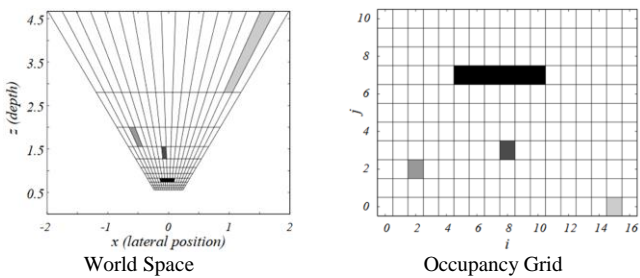

Figure 3. Column/Disparity Occupancy Grid from [11].

Similar to [11] we define the column/disparity occupancy grid. Contribution of the $v$ component of the measurement $m_k$ can be ignored as it contributes only to the height component of the 3 D point/measurement. Once we project the 3D points onto the ground plane the height component of the measurement is lost so we do not consider it here. So our measurement component then becomes of the form $m_k = (u_k, 0, d_k)$. Say cell $(i, j)$ corresponds to the coordinate $(u_{ij}, d_{ij})$. We define the occupancy likelihood for the cell as

$$L_{ij}(m_k) = G_{m_k}((u_{ij} - u_k, 0, d_{ij} - d_k)) \quad (3)$$

Above equation gives us the occupancy likelihood of a measurement. Summing this over all measurements for a particular cell using equation (2) we get the occupancy likelihood of the grid $(i, j)$.

### C. Segmentation of Free Space

As opposed to [11], we do a segmentation of the free space from the non-free space using a thresholding based approach. After computing the Occupancy Likelihood of the occupancy grid, we traverse each column of the 2D array one by one and as soon as we find a particular threshold met in some row we claim that to be the first occupied grid and behind that everything is considered occupied as everything behind the object is occluded by the object. Figure 3 illustrates this approach. Thresholding based segmentation is faster than the dynamic programming approach used in [11], hence can be implemented easily in real time. Thresholding based segmentation leads to a solution without any spatial or temporal smoothness and non-global optimization. Also the segmentation results highly depend on the input threshold set for the likelihood. Another segmentation approach was also explored which makes counts the transitions as shown in figure 6.5 and 6.11 (black -> white->black -> white). The first white pixel of the second transition from black to white region is considered as the segmentation boundary in each column. This also proves to be an efficient measure in finding the segmentation.

| 43 | 241 | 24 | 31 | 2 | 343 | 45 | 21 |
|---|---|---|---|---|---|---|---|
| 442 | **56.2** | 12 | 32 | 21 | 45 | 13 | 54 |
| 121 | 22 | 112 | 45 | 454 | 32 | 13 | 1 |
| **45.3** | 5.2 | 234 | **78** | 223 | **45** | **90** | 4 |
| 2.43 | 1.4 | **112** | 4 | **56** | 6 | 11 | **67** |
| 0.54 | 4.3 | 23 | 3 | 34 | 14 | 21 | 5 |
| 0 | 2.9 | 0.3 | 7 | 12 | 17 | 23 | 2 |
| 0 | 1.5 | 4 | 4 | 4 | 8 | 2 | 5 |

Figure 4. Thresholding based segmentation on a sample occupancy grid with computed likelihoods, here we set the threshold as 40. Grey colour indicates occupied, white indicates free space. Arrow indicates direction of search along each column.

### D. Mapping Consecutive Occupancy grids

After computing free space from a pair of stereo images and computing the top view representation, we now need to build a map of the environment by iterating over the sequence of images. We deal with mapping with dynamic objects in the section IV. Here we describe mapping without dynamic objects which we extend in the next section for dynamic object mapping too.

An affine transformation is computed between the 2 occupancy grids computed in frame $i$ and frame $i + 1$. A global coordinate system is defined and a local coordinate system is defined for each occupancy grid. A transform of the local coordinate system w.r.t. the global coordinate system computes the relative positions of the current grids in frame $i$ and maps them to frame $i + 1$. The new grids observed in frame $i + 1$ are updated in the global coordinate system to build the map. This process is repeated for the subsequent frames to build a continuous map of the world.

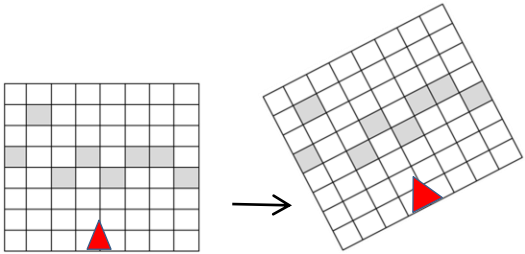

Figure 5. Red triangle indicates the ego car pose. An Affine Transformation between the 2 occupancy grids is computed w.r.t. to a global coordinate system

IV. DEALING WITH DYANMIC OBJECTS FOR MAPPING

One of the most challenging aspects for mapping is addressing issue of dealing with dynamic objects in a scene. One needs to distinguish dynamic objects from fixed obstacles/background in the scene in order to build a map because we do not want to keep track of dynamic object on the final map built. The way we propose to address this issue is inspired from the topic of segmenting/tracking motion of dynamic objects from the monocular image sequences taken from a moving camera. In our case the moving camera is the one mounted on the car. We want to segment/divide/cluster the flow of the foreground moving objects and the background. We formulate this as two-view based motion segmentation problem.

*A. Learning a model from keypoints to segment dynamic object motion from background motion using RANSAC algorithm.*

We first find key point descriptors in the frame $i$ and frame $i + 1$. For key-point extraction we use the maximally stable extremal regions (MSER) detector proposed by [27] and Harris Features. After extraction of the key-points, we extract descriptors around the key points. For MSER regions we extract SURF descriptor [29] around the key points and for Harris Points we use the FREAK [30] method to extract features. After feature extraction we match the features extracted from frame $i$ to those in frame $i + 1$. Now we have a set of points $P_i$ from frame $i$ and the corresponding matched points $P_{i+1}$ in frame $i + 1$. We now want to learn a model for the perspective background motion of the camera. By doing this we know the likely motion of the background. All points which do not obey the learnt model are considered to be from the dynamic moving object. Figure 5 gives an overview of our approach for detecting dynamic object motion. We now want to learn a function $f$ which takes as input the key-point coordinates and the flow of the key-point and predicts the magnitude of the flow $\psi$. We define a cubic model function $f$ whose parameters need to be estimated to learn the model:

$$f(u, v, \psi) = \psi = c_1 u^3 + c_2 v^3 + c_3 u^2 v + c_4 u v^2 \quad (4)$$
$$+ c_5 u^2 + c_6 v^2 + c_7 u v + c_8 u$$
$$+ c_9 v + c_{10}$$

$$\psi = \sqrt{f_u^2 + f_v^2} \quad (5)$$

$$f_u = u_i - u_{i-1} \quad (6)$$

$$f_v = v_i - v_{i-1} \quad (7)$$

where $u, v$ are the image coordinates of the key-point; $c_1, c_2, \ldots c_{10}$ are the model parameters to be found. $f_u, f_v$ are the flow in $u$ and $v$ directions respectively and $\psi$ is the magnitude of the flow. $u_i, u_{i-1}, v_i, v_{i-1}$ are the image coordinates in frame $i$ and $i - 1$. Now we want to find the function curve $f$ such that all key-points on the back ground fit it well and the foreground dynamic objects do not fit it well and result in a high value. We use the random sampling and consensus (RANSAC) algorithm [28] for doing the same. Since RANSAC is a robust method and outliers have no effect on the model which is learnt. Also it is assumed that the number of background key-points detected are much more than the dynamic objects which is true for the image sequences in the KITTI dataset. The outlier points would belong to the dynamic objects. Based on the selection of our key-point extracted, we ensure that the number of chosen points is more on the background than the foreground. The function proposed in (4) basically learns the perspective motion of the camera between 2 consecutive frames. We also use a linear and quadratic function to estimate the flow and get good results using them too, however in some cases the cubic function models the background motion better than the quadratic function. Polynomials of higher degree {4,5,6} were also tried and gave good results too but they were computationally expense as the number of parameters to be estimated increases as the polynomial's degree increases. In this paper we stick to using linear, quadratic and cubic functions

Some valid assumptions that we consider here are
- Dynamic objects have a flow different from the background.
- Number of key-points in the background is much more than the key-points in the dynamic moving objects.
- Dynamic Moving object has a MSER or a Harris Point.

It should be noted that all these assumptions hold true for the KITTI benchmark.

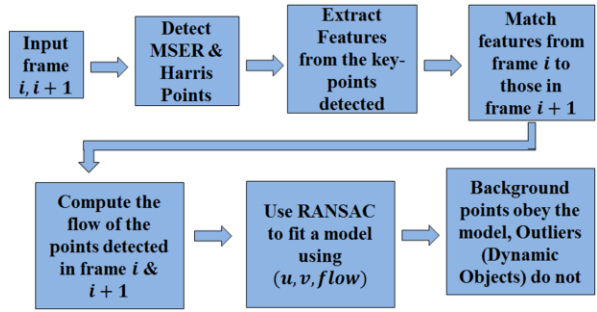

Figure 6.  General Pipeline for detection of Dynamic Objects.

This proposed model is learnt for every two consecutive frames, so even when the car stops the back ground motion is zero and the dynamic motion would be easily segmented out as they would have a different motion from the background motion model. The proposed model assumes that dynamic objects must have a different motion from the background motion at the dynamic object key-point. As long as this condition holds true the proposed model successfully detects dynamic motion. It should be noted here if a dynamic object is stationary and the car is also stationary the model would consider the dynamic object as a background point.

### B. Using Prior Knowledge to detect

In this subsection we present an alternative approach to solve dynamic object detection. Using a prior assumption that only cars, humans and 2 wheelers (bikes) can be on the road; detection of these can help in mapping in environments with dynamic objects too. If a car is parked one can simply use the approach discussed in the previous section to find out that it has the same flow as the background and hence is classified as a static object. One can train highly efficient object classifiers using CNN based techniques [31] or using traditional hand crafted approaches. Classifiers/Detectors specific to cars [32], humans [33], [34] and bikes can be trained. This subsection is proposed as a possible solution to the problem; it has not been implemented.

### C. Mapping with Dynamic Objects

As was discussed in Section III, mapping without dynamic objects can be effectively done by computing a transformation between the 2 subsequent occupancy grids maps. With dynamic objects we propose to do the mapping as follows. Once we detect the key-points of the dynamic objects we descrease the occupancy likelihood of the cell to which the key-point belongs and designate it be unknown as the dynamic object may be occluding a fixed obstacle. In the subsequent frames after dynamic object moves out of the way if the occupancy likelihood of that particular cell increases behind the region of the dynamic object we say it is occupied else not. If the dynamic object does not get out of the way in time in the subsequent frames and we are unable to see behind the object we conclude the region to be unknown in the map.

### D. Mapping with Dynamic Objects

We now present the limitations of our proposed approach. Our approach relies on fitting a polynomial model function to segment background motion which is currently not real time implementable. However after incorporating some optimization in the code it would be able to run in real time. The presented approach cannot handle cases where mirror is present which leads to detection of dynamic objects in the mirror too. Our model treats shadows of moving objects also to be a part of the dynamic object.

## V.  EXPERIMENTAL EVALUATION

In this section we present the results obtained from our proposed approach. We divide this section into 2 parts. First we show the results obtained for free space computation using occupancy grids. Next we show the detection of dynamic objects as proposed in Section IV. For experimental evaluation we use sequences of images obtained from the KITTI benchmark dataset.

### A. Computing Free Space based on Occupancy Grids

For computing free space using occupancy grids, we first compute the disparity by using the displets approach of [22]. Then we compute the occupancy likelihoods, and finally use the thresholding based segmentation to compute the free space. We show the results obtained by our approach in figure 6. For mapping we compute a transform between figure 6.6/6.12 and their subsequent images to generate a top view map.

### B. Detection of Dynamic Objects

Here we detect dynamic motion based on the approach proposed in Section V. We use a RANSAC based model fitting to find out the outliers which are the dynamic objects in our case. It is assumed that the number of key-points of the background is more than those in the moving objects. We show the result of detection and segmentation of the dynamic objects in this section. Figure 7 shows the results of our approach. We tried polynomials upto degree 3 – linear, quadratic and cubic models to be learnt by RANSAC. We choose 40% of the total detected points and set the number of iterations to be 20 in RANSAC for the dynamic objects detection. The results presented here use the linear model learnt by RANSAC. For predicting the magnitude of flow we set the coordinate axis to be in the center of the image as shown in Figure 8. It should be noted a RANSAC based model was also trained to model the orientation of the flow, however it was observed that the model was not effective enough for far away objects. A future extension of the work would include combining the model to estimate the magnitude and the orientation of the flow together to perform the segmentation of the dynamic object.

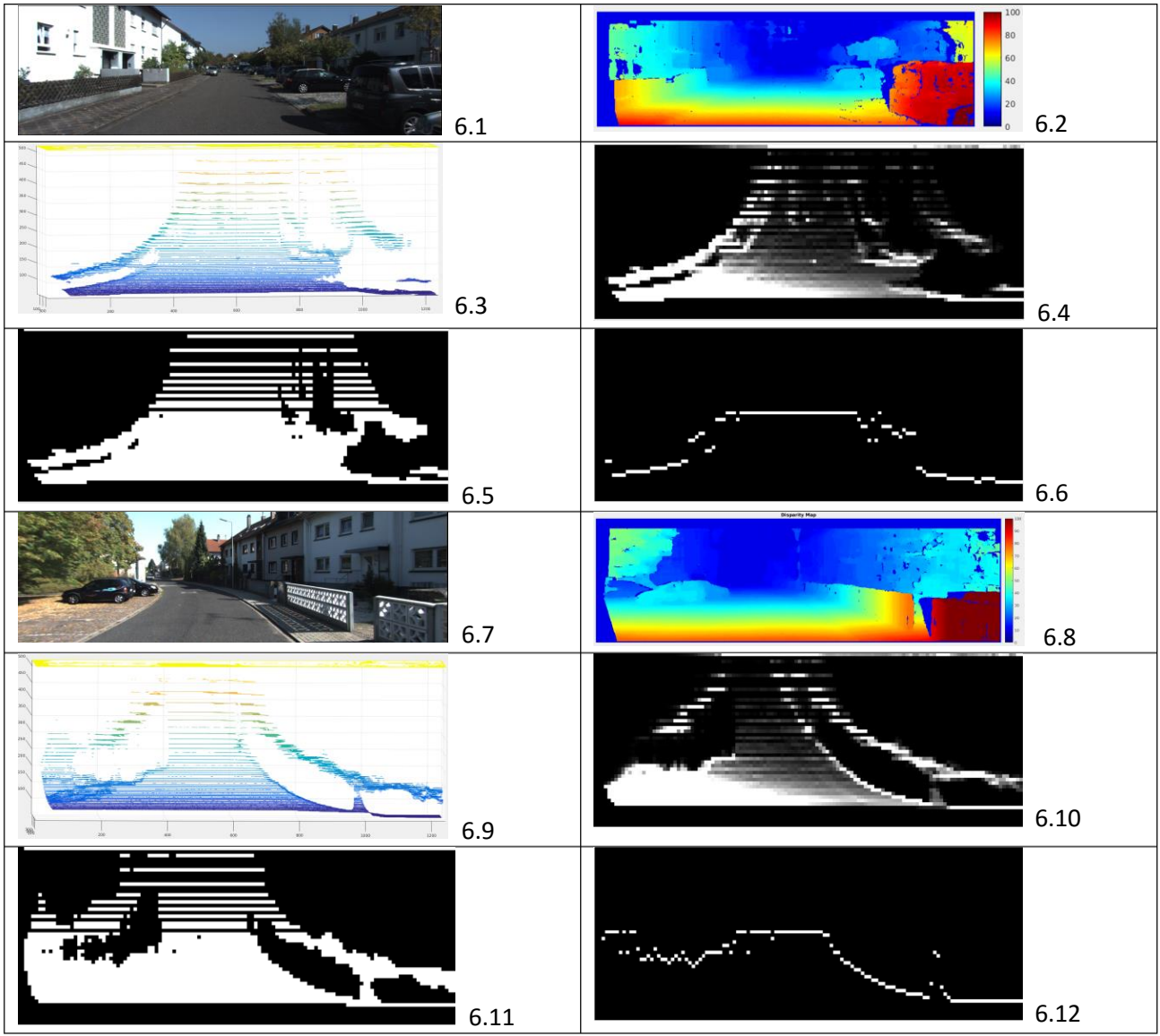

Figure 7. 6.1/6.7 input images, 6.2/6.8 Disparity Maps, 6.3/6.9 Top view of the point cloud, 6.4/6.10 Occupancy Likelihood intensity, 6.5/6.11 Binary occupancy grid, 6.6/6.12 Segmentation of the Occupancy grid to indicate Free Space and Non-Free space (bird's eye view)

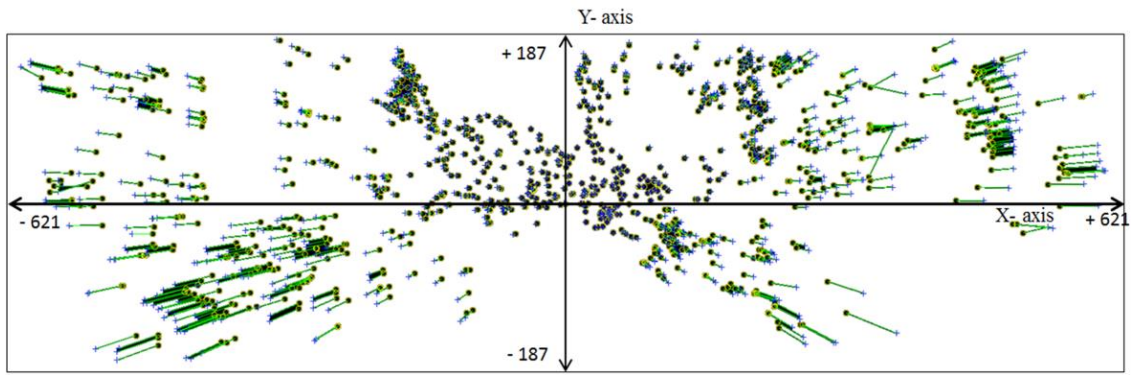

Figure 8. Based on the Flow vectors and image coordinates (u,v) = ((x,y) axis here) of the background feature points, the models parameters are estimated in a RANSAC fasion.

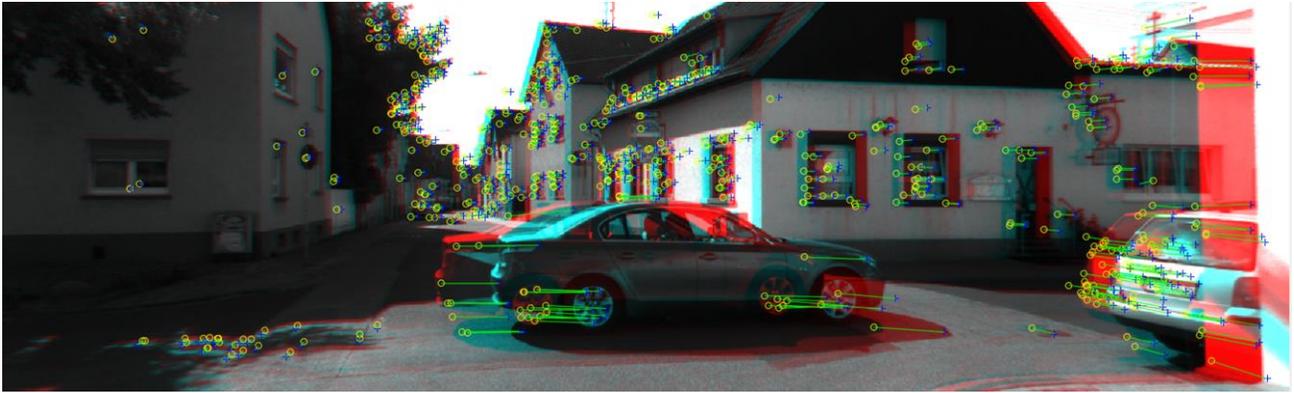

(i)

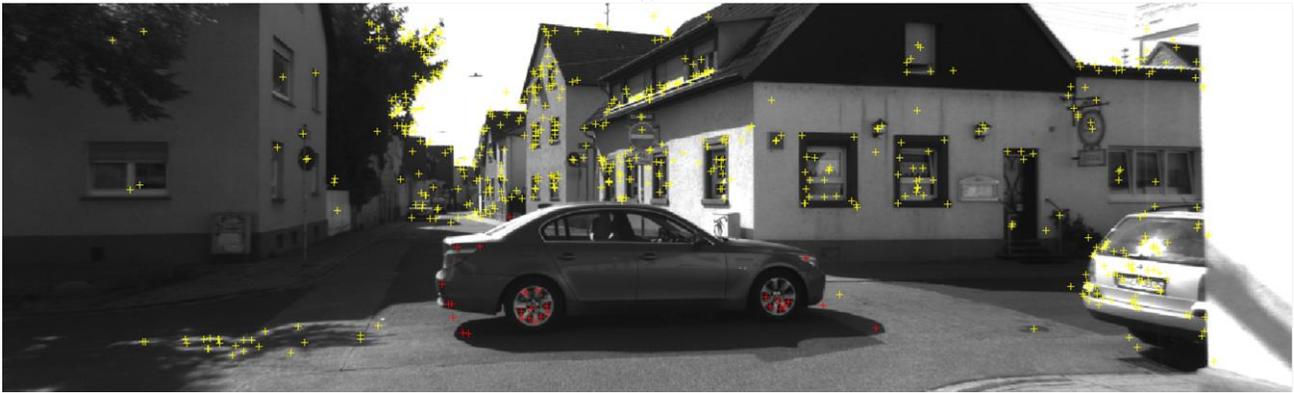

(ii)

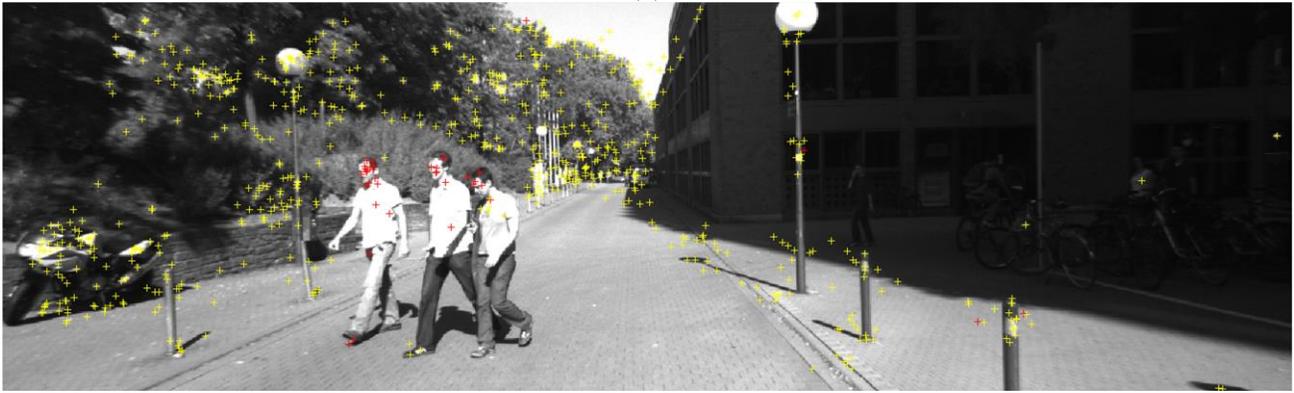

(iii)

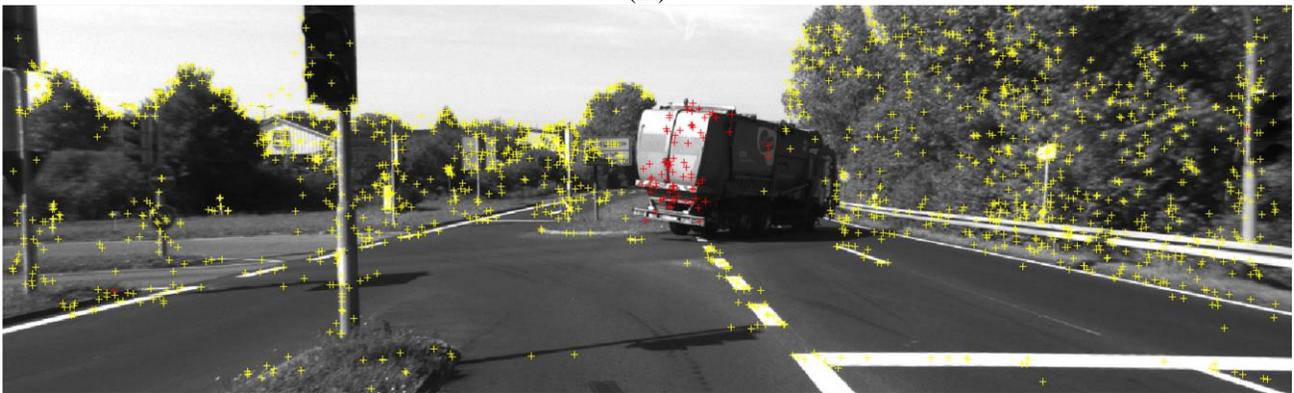

(iv)

Figure 9. Matching Harris and MSER Features in 2 consecutive frames (i); Segmentation of Dynamic Motion Keypoints using the RANSAC based model (ii,iii,iv)

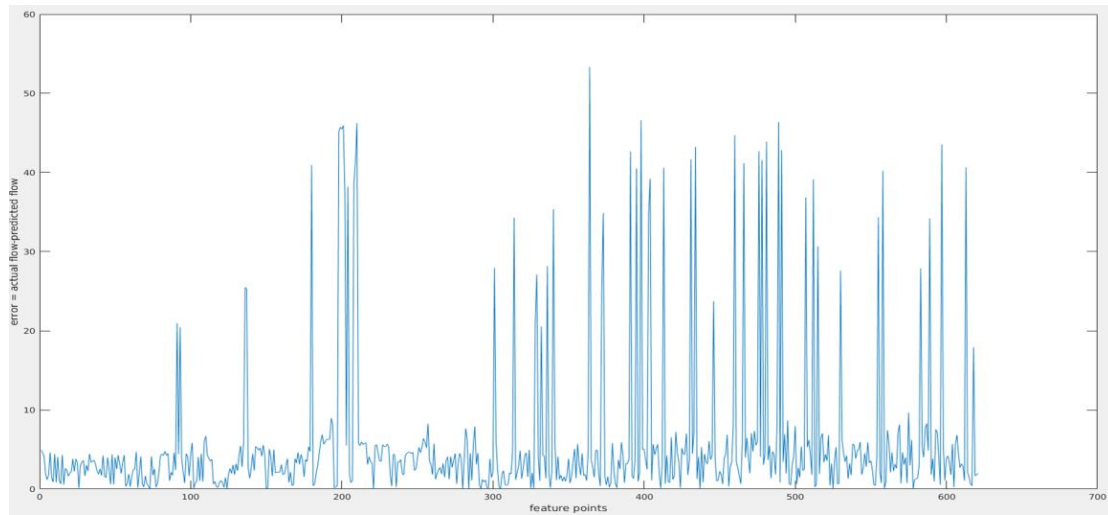

Figure 10. Plot showing error in estimated flow of Feature Points using the RANSAC model. Points having large error are the outliers (dynamic object points) to the model.

Our model is able to handle dynamic object with different flow than the background. A limitation of the presented approach includes that of inability to detect dynamic object that are far away in the scene. Objects that are far away in the scene are not detected. An interesting thing about our approach is one can train a Gaussian mixture model to segment the dynamic motion of different objects. So an extended approach of this work could be fitting Gaussian mixture models to distinguish between motions of different dynamic objects. By training a Gaussian mixture model the approach can also be applied to segment articulated motion.

## VI. IMPLEMENTATION

The code was implemented in matlab and tested on a sequence of images obtained from the KITTI dataset. We make the code publically available at the following link:
https://github.com/raghavendersahdev/Free-Space

Currently we have not written the code for updating the map with dynamic objects; it contains the code for computing free space from a single image using stochastic occupancy grids and detection/segmentation of motion of dynamic object by the approach discussed in Section V.

## VII. CONCLUSION

In this paper we presented a method to compute drivable free space estimation and presented an approach to build a top view map of the environment from a sequence of images. We then proposed a novel RANSAC based algorithm for detection of dynamic motion. Finally we tested our proposed approaches against the KITTI benchmark dataset and got good results on it. We also discussed some ideas about updating the map in the presence of dynamic objects.

Possible future work includes exploring more robust models to be optimized by RANSAC to learn more efficient motion models for the background leading to faster detection of dynamic objects. Some possible future work includes building a robust method to integrate this approach for building maps in the presence of dynamic objects. Integrating information from dynamic motion segmentation needs to be incorporated for building high quality free space maps as seen from a birds eye. Once a map of the environment is build one could extend this work to autonomous navigation of vehicles. Same approach can also be used for navigation of mobile robots in a dynamic environment. Some of the immediate future work following this work is to test the proposed approach of free space estimation and dynamic object detection on the indoor environment dataset built by Sahdev and Tsotsos [39] with the introduction of dynamic objects in the scenes.

## ACKNOWLEDGMENT

The author would like to thank Raquel Urtasun for the helpful discussions during the course of this project.